# Explainable AI-Guided Efficient Approximate DNN Generation for Multi-Pod Systolic Arrays

Ayesha Siddique, Khurram Khalil, Khaza Anuarul Hoque
*Department of Electrical Engineering and Computer Science*
*University of Missouri, Columbia, MO, USA*
{ayesha.siddique, khurram.khalil, hoquek}@missouri.edu

*Abstract*—Approximate deep neural networks (AxDNNs) are promising for enhancing energy efficiency in real-world devices. One of the key contributors behind this enhanced energy efficiency in AxDNNs is the use of approximate multipliers. Unfortunately, the simulation of approximate multipliers does not usually scale well on CPUs and GPUs. As a consequence, this *slows down* the overall simulation of AxDNNs aimed at identifying the appropriate approximate multipliers to achieve high energy efficiency with a minimum accuracy loss. To address this problem, we present a novel XAI-Gen methodology, which leverages the analytical model of the emerging hardware accelerator (e.g., Google TPU v4) and explainable artificial intelligence (XAI) to precisely identify the non-critical layers for approximation and quickly discover the appropriate approximate multipliers for AxDNN layers. Our results show that XAI-Gen achieves up to $7\times$ lower energy consumption with only 1–2% accuracy loss. We also showcase the effectiveness of the XAI-Gen approach through a neural architecture search (XAI-NAS) case study. Interestingly, XAI-NAS achieves 40% higher energy efficiency with up to $5\times$ less execution time when compared to the state-of-the-art NAS methods for generating AxDNNs.

## I. INTRODUCTION

Approximate deep neural networks (AxDNNs) have made a breakthrough in edge analytics. This achievement is primarily attributed to approximate multipliers, which can help achieve energy savings exceeding 28% [1], [2]. Nevertheless, to fully harness the potential of approximate multipliers in AxDNNs, an exhaustive search to identify the most suitable layer-wise combination of approximate multipliers is indispensable [1]. Indeed, such an exhaustive search can help achieve the best trade-off between accuracy and energy efficiency in AxDNNs. However, the simulation of approximate multipliers does not scale well on CPUs and GPUs [3], which slows the overall search process and makes such an exhaustive search impractical for large designs. Researchers have tried to propose different solutions to solve the above-mentioned problem. Since approximation errors originating from approximate multipliers exhibit a Gaussian distribution, several studies exploit Gaussian noise injection to identify resilient DNN layers [4] and then apply approximate multipliers solely to these resilient

This material is based upon work supported by the National Science Foundation (NSF) under Award Numbers: CCF-2323819. Any opinions, findings, conclusions, or recommendations expressed in this publication are those of the authors and do not necessarily reflect the views of the NSF.

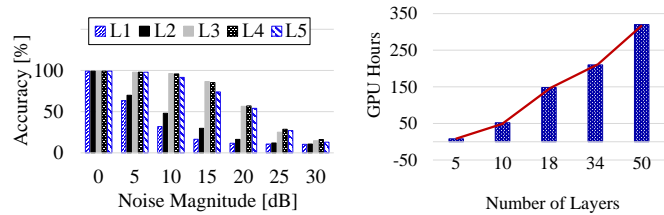

Fig. 1: Approximation error resilience of layers L1 to L5 in Lenet-5 using White Gaussian Noise and execution time of AxDNNs. Noise magnitude of 0 dB denotes Lenet-5 without approx. noise.

layers to accelerate AxDNN simulations. Note that selecting the appropriate Gaussian noise magnitude and approximation multipliers involves a trial-and-error method, which can be prone to human error. As an alternative approach, a mathematical model [2] estimates the magnitude of approximation noise in approximate multipliers of AxDNNs. However, it relies on the input values obtained from all convolutional layers, which may change in real-time applications. Also, when such a model is evaluated with a uniform distribution of activations, the resulting noise magnitude differs from the noise magnitude estimated for approximation multipliers in a pre-trained model. Other strategies involve exploiting the protobuf format [5] and Cartesian Genetic Programming (CGP) [6] to find the most suitable approximate multiplier serving multiple AxDNN layers. Though all these works are important, there are two major gaps: (i) most of these works use the same type of approximate multipliers in a layer and do not explore the possibility of using heterogeneous approximate multipliers in each layer for improved accuracy and energy consumption tradeoff, and (ii) with all these proposed approaches, it still takes a considerable amount of time to AxDNNs even for small datasets.

Recently, explainable artificial intelligence (XAI) has gained a lot of attention in different application domains [7]. XAI aims to provide deep insight into the factors that contribute significantly to the decision-making process in deep neural networks. For instance, XAI techniques reveal the importance of neurons and weights in a neural network [8]. By analyzing this information, the designers *precisely* skip the neurons, which have minimal impact on the neural network's accuracy

[7], [9]. In contrast, important neurons or weights that substantially influence accuracy are preserved. XAI techniques have also been utilized to speed up the fault simulations [10]. *Surprisingly, the idea of applying XAI to guide the search method for faster generation of AxDNNs is still unexplored.*

### A. Motivational Case Study and Key Observations

To highlight the limitations of state-of-the-art works, we present a motivational case study focused on (i) the impact of Gaussian noise injection on AxDNNs using a trial and error method, (ii) the implications of selecting approximate multipliers based on the estimated magnitude of Gaussian noise using a mathematical model [2], and (iii) the simulation time required for extensive approximation analysis. Our experiments employ a LeNet-5 architecture with 3 convolutional and 2 fully connected layers. We inject the Gaussian noise of varying noise magnitudes in these layers to estimate their resilience to approximation errors. Fig. 1 shows our results for the case study, from which we make the following key observations:

1) Selecting the noise magnitude through the **trial and error method** results in comparatively distinguishable noise resilience in layers L1 and L2 at 10 dB and 15 dB. Other layers L4 and L5 have a slight difference in accuracy drop at 15 dB only. The accuracy drop is not discernible across layers L3 to L5 for other noise magnitudes. This makes sorting the resilient layers with an appropriate noise magnitude challenging (see Fig. 1a).
2) Injecting a Gaussian noise with magnitude 0.0736, provided in [2] for a **uniform distribution**, causes an accuracy loss of 10%. However, the corresponding approximate multiplier, with a mean average error (MAE) of 5.09%, results in a 60% accuracy loss. Thus, [2] is not a reliable method for selecting approximate multipliers in neural networks except DeepCaps [11].
3) An **extensive approximation analysis** takes 9 GPU hours to find the most suitable configuration from 3K combinations of 5 approximate multipliers in a 5-layered AxDNN. Increasing the AxDNN complexity by $10\times$ results in a $34\times$ surge in the GPU hours required (see Fig. 1b).

These observations raise a key research question: *how to expedite the AxDNN generation time by precisely identifying the most suitable configuration of heterogeneous approximate multipliers for each layer to achieve the best accuracy-energy tradeoffs?*

### B. Novel Contributions.

To address the above research challenge, we propose the novel e**X**plainable **A**rtificial **I**ntelligence (XAI) guided AxDNN **Gen**eration (**XAI-Gen**) method, which uniquely leverages neuron sensitivity analysis for heterogeneous approximate multiplier selection. Our novel contributions include:
1) An analytical model of the layers and operations of neural networks when mapped to emerging multi-pod systolic array accelerators (e.g., Google TPU v4 [12]) and their approximate counterparts. Running AxDNN simulations using this analytical model helps in architectural model flexibility and fast hardware efficiency estimation of AxDNNs **(see Section III)**.
2) A novel AxDNN generation method that leverages XAI and the analytical model to precisely identify the non-critical layers for approximation and quickly discover the appropriate approximate multipliers for AxDNN layers for achiving higher energy efficiency at the cost of minimal accuracy loss **(see Section IV)**.
3) A comprehensive performance analysis of XAI-Gen generated AxDNNs regarding accuracy-energy tradeoffs for approximate multi-pod systolic array accelerators. We also present the results for an XAI-Gen powered neural architecture search (NAS) case study. **(see Section V)**.

**Key Results.** To demonstrate the effectiveness of our proposed XAI-Gen method, we evaluate it on various CNN architectures using three datasets: MNIST [13], CIFAR10 [14], and ImageNet. We compare our approach with state-of-the-art approximation techniques, including pruning and quantization methods. We generate AxDNN architectures by replacing the accurate multipliers with approximate multipliers from Evoapprox8b [15] library. We employ LeNet-5 for MNIST and ResNet-50 for ImageNet. We also use ResNet-18 and ResNet-34 for CIFAR-10. Our results show that XAI-Gen achieves up to $7\times$ lower energy consumption with only 1–2% accuracy loss, while significantly reducing the search time for optimal approximate configurations compared to existing methods. We also demonstrate the effectiveness of our proposed XAI-Gen in NAS (XAI-NAS). XAI-NAS achieves 40% higher energy efficiency with up to $5\times$ less execution time as compared to state-of-the-art NAS methods [5], [6] for generating AxDNNs. To the best of our knowledge, this is the first work to propose AxDNN generation using XAI.

The remainder of this paper is structured as follows: Section II discusses the state-of-the-art works on AxDNNs, XAI, and multi-pod systolic array accelerators. Sections III presents the analytical model of multi-pod systolic array accelerator. Sections IV elucidates the proposed XAI-Gen design approach and methodology, and presents an application of XAI-NAS. Section V discusses the results obtained from XAI-Gen and XAI-NAS evaluation. Finally, Section VI concludes the paper.

## II. BACKGROUND: MULTI-POD SYSTOLIC ARRAYS

This section provides a brief overview of state-of-the-art AxDNNs, XAI, and multi-pod systolic arrays.

### A. Approximate Deep Neural Networks

AxDNNs relax the abstraction of absolute precision by employing approximate multipliers either uniformly across all layers or in a non-uniform structure [16], [17]. In the non-uniform approach, AxDNN layers with higher error resilience integrate approximate multipliers with a higher approximation for more potential energy savings. In contrast, layers with lower error resilience undergo precise computations to maintain acceptable accuracy [5]. Hanif et al. investigated the

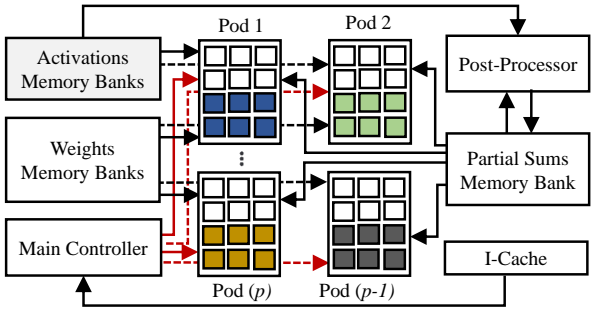

Fig. 2: Multi-pod approximate systolic array accelerator

Gaussian distribution of approximation errors in approximate multipliers, providing the opportunity to speed up such simulations [18]. The authors identified the resilient and non-resilient layers through the Gaussian noise injection prior to integrating approximate multipliers in AxDNNs. However, the noise magnitude is selected randomly and therefore, does not provide a reliable estimate of layer resilience. Marchisio et al. determined the noise magnitude by modeling the approximation errors from approximate multipliers [2]. However, this model relies on the range of input values from all convolutional layers, limiting its applicability in real-time scenarios due to the computational overhead of analyzing all layers.

### B. Explainable Artificial Intelligence

Neural networks are highly complex and often considered black-box models because their inner workings are not readily interpretable by humans. XAI bridges this gap by providing insights, explanations, and interpretability to the functioning of neural networks. A comprehensive survey of existing XAI techniques can be found in [8], most of which are related to explaining deep learning models, known as a post-hoc explanation. Most of the algorithms used in post-hoc explanations can be grouped into gradient and perturbation-based approaches. Several tools are available for post-hoc explanations through calculating feature or neuron importance in neural networks, including LIME, Captum, and SHAPly [19]–[21] etc. The Captum tool, which we use in this paper, supports the identification of neuron importance in DNN models. This capability distinguishes it from LIME and SHAPly, which primarily focus on identifying feature importance in DNN models.

### C. Multi-pod approximate systolic array accelerator

A monolithic systolic array employs a two-dimensional grid of multiply-and-accumulate (MAC) units, which process the data from the neural network layers in a parallel fashion. However, such a grid may not be fully utilized when confronted with small layers. Indeed, a significant performance disparity among layers may lead to wait times for processing the computationally intensive layers. Modern multi-pod systolic array accelerators like Google TPU v3 and v4 [12] address these challenges by clustering a few coarse-grain systolic array pods connected via shared memory on a single die. They partition the weight matrix $W$ into $r \times c$ tiles matching the rows $r$ and columns $c$ of the systolic array pods. Yet, the activation matrix $X$ is partitioned with a size of $r$ along the second dimension only. Yuzuguler et al. proposed full data-level parallelism through partitioning $X$ into $r \times r$ tiles [22]. Specifically, the weight and activation matrices are mapped to the systolic array pods such that their MAC units perform $x_{ij} * w_{jk} + y_{imk} = y_{ijk}$ in every time slice. The $x_{ij}$ and $w_{jk}$ are tiles from $X$ and $W$, $y_{imk}$ is the propagating partial sum, and $y_{ijk}$ is the resulting partial sum. $x_{ij}$ and $y_{imk}$ propagate through consecutive MAC units along the rows and columns of the available systolic pods, with some offset, at the earliest available time. The resulting partial sums $y_{ijk}$ are processed through activation functions, generating the final outputs $\sigma_{ij}$ using post-processors. Fig. 2 illustrates the structure of a multi-pod approximate systolic array accelerator.

### III. ANALYTICAL MODEL OF MULTI-POD SYSTOLIC ARRAYS

To evaluate the energy efficiency of neural network models on specialized hardware accelerators like multi-pod systolic arrays, it is crucial to understand their underlying hardware characteristics. These hardware characteristics include time period $T$, power consumed $P_{pod}$ by the $m$ number of MAC units in a systolic-array pod, and energy required per memory access $E_M$. Knowing these details enables the analytical modeling of energy consumption. In this paper, we consider an approximate multi-pod systolic array accelerator. We assume that the lower halves of the systolic-array pods have approximate MAC units, as shown in Fig. 2. For example, an approximate variant of Google TPU v4 can have up to four distinct approximate systolic array pods, each having a different approximate multiplier in the lower half. Mapping heterogeneous AxDNN layers $l$ to such approximate accelerators improves the utilization of energy-efficient computational resources. We assume a total number of weights $n_w$ and a number of feature maps $f_l$ to be multiplied by the same weight in a layer and the number of clock cycles $c_l$ needed to process a layer. Given these parameters, the number of groups of weights loaded into each half of such an accelerator can be calculated as $w_{p/2} = \frac{2n_w}{(m \times p)}$. Also, the number of memory accesses $N_m$ can be determined as follows:

$$N_m = \begin{cases} r \cdot c & \text{if } (r \cdot c) < (\frac{m}{2}) \\ (\frac{m}{2}) \cdot max(f_l - ((\frac{r \cdot c}{2}) - 1), 1) & \text{if } (\frac{m}{2}) < (r \cdot c) < m \\ (m) \cdot max(f_l - (r \cdot c - 1), 1) & \text{otherwise} \end{cases} \quad (1)$$

The above-mentioned $w_p$ and $N_m$ can be used to estimate the energy consumption ($E_a$) are given as follows:

$$E_a = \lceil \frac{N_m}{\mathcal{R}} \rceil \cdot E_M + \sum_{l \in L} c_l \cdot T \cdot P_{pod} \quad (2)$$

where $\mathcal{R}$ and $L$ represent a number of rows per memory bank and a total number of layers in the AxDNN model, respectively. The number of clock cycles is given as $c_l$ as $c_l = w_p$. The $P_{pod}$ is calculated as $(m/2).P_{mac} + (m/2).P_{xmac}$. The $P_{mac}$ and $P_{xmac}$ denote the power consumed by accurate and approximate MAC units, respectively. The analytical model of this accelerator has been validated by comparing the results with the hardware implementation of the multi-pod systolic array accelerator in [22]. Our analytical model provides an accurate estimate of latency and energy consumption.

## IV. XAI-GEN: PROPOSED METHODOLOGY

This section discusses our proposed approach and methodology for XAI-guided AxDNN generation, and a case study.

### A. XAI Guided Approximate Computing

We use XAI to analyze gradients within a neural network to assess the importance of individual neurons and entire layers. These insights are valuable for making informed decisions regarding selecting appropriate approximate multipliers in AxDNN layers. When analyzing the importance of neurons, it is crucial to understand their contributions relative to a baseline. Let us consider an AxDNN classifier $\mathcal{F} : \mathcal{R}^n \rightarrow [0, 1]$, an input $x = (x_1, ..., x_n)$, and a baseline input $x'$ i.e., black image. The relative importance of an input $x$ is given as $(a_1, ..., a_n) \in \mathcal{R}^n$, where $a_i$ can be interpreted as the attributions of $x_i$ to the output $\mathcal{F}(x)$. This attribution can be used for intermediate neurons instead of input $x$. The relative importance of a neuron $y$, also known as neuron conductance $\mathcal{C}$ [23], for the attribution to an input variable $i$ is calculated as:

$$\mathcal{C}_y(x) = \sum_i (x_i - x_i') \cdot \int_{\gamma=0}^{1} \frac{\partial \mathcal{F}(x' + \gamma(x - x'))}{\partial y} \cdot \frac{\partial y}{\partial x_i} d\gamma, \quad (3)$$

where $\gamma$ is a scaling factor for the difference between $x$ and $x'$. As shown in Fig. 3, the neuron conductance $c_i$ for each $i^{th}$ neuron in a layer is calculated by analyzing the gradients using the forward and backward passes. The neuron conductance for each neuron in a layer can be summed to find the importance of the entire layer $z$. This information is indeed helpful for precisely classifying the network layers into different categories ranging from critical to non-critical and strategically applying approximate multipliers $m$ to them. For example, the critical layers significantly affect the neural network's performance; therefore, they can be allocated more resources to ensure their proper functioning. They can have either no or minimal approximation, with approximate multipliers having low approximation error. On the other hand, non-critical layers, being less influential, can potentially be subject to aggressive approximation without causing substantial degradation in accuracy. They can have approximate multipliers that exhibit relatively higher approximation errors.

### B. Proposed Methodology

Our proposed methodology (see Fig. 4) accelerates the AxDNN generation by leveraging XAI and the analytical model of the multi-pod systolic array accelerator. Leveraging XAI enables precise identification of the non-critical layers for approximation and quickly finding the appropriate approximate multipliers for AxDNN layers. The proposed methodology comprises five key steps: estimating the neuron importance, determining layer importance, XAI-guided neuron skipping and approximation, estimating energy consumption, and analyzing the accuracy-energy tradeoffs. Algorithm 1 delineates the steps involved in generating XAI-guided AxDNNs.

**Algorithm 1:** XAI-Gen: XAI-guided AxDNN generation

**Inputs:** Number of layers in trained model $\theta$: $L$;
Number of neurons in layers: $Y_l$;
Set of approximate multipliers: $m$;
Step size: $\delta$; Initial conductance threshold: $t_c$;
Skipping threshold: $t_p$; Quality constraint: $Q_c$
Energy constraint: $E_c$; Quantization scale: $q$
Error magnitude of multipliers: $M_g$; Dataset: $\mathcal{D}$

**Outputs:** AxDNN model: $\theta^*$; $E_l$; $Q_l$

1: $R^* \leftarrow \{m_0, m_1, ....ml\}$
   // Initialize a list of accurate multipliers for each layer
2: **for each** $l$ in $L$ **do**
3:   **for each** $y$ in $Y_l$ **do**
4:     $IG(x) \leftarrow \int_{\gamma=0}^{1} \frac{\partial F(x' + \gamma(x-x'))}{\partial y} \cdot \frac{\partial y}{\partial x_i} d\gamma$
       // Integrated gradients for each neuron
5:     $\mathcal{C}_{y,l}(x) \leftarrow \sum_i (x_i - x_i') \cdot IG(x)$
       // Neuron conductance for each neuron
6:   **end for**
7:   $z_l \leftarrow \sum_j \mathcal{C}_{y,l}(x)$
     // Aggregated neuron conductance for each layer
8: **end for**
9: $Zval_l \leftarrow AscendingSort(z_l, L)$
   // Sort $z_l$ from least to most important layer
10: **while** (traversing layers with $Zval_l$) **do**
11:   $(m_l^*, R^*) \leftarrow SelectMult(Zval_l, t_c, M_g, m)$
      // Select the approximate multiplier $m_l^*$, and
      // update the list $R^*$ for layer $l$
12:   $n_p \leftarrow (\mathcal{C}_{y,l}(x), t_p, \theta)$
      // Indices of least important neurons
13:   $\theta_p \leftarrow (n_p, \theta_q)$
      // Skipping the least important neurons
14:   $\theta_q \leftarrow Quantize(q, \theta_p)$
      // Quantize models
15:   $\theta^* \leftarrow Generate(R^*, \theta_q)$
      // Generate AxDNN models
16:   $(E_l, Q_l) \leftarrow Evaluate(R^*, \mathcal{S}, \mathcal{D}, \theta^*)$
      // Evaluate energy efficiency and quality using the
      // analytical model $\mathcal{S}$ and model $\theta$
17:   **if** $(E_l > E_c)\&(Q_l < Q_c)$ **then**
18:     **break**
19:   **end if**
20:   Commit $t_c$ and $m_l^*$ in layer $l$
21:   Reduce $t_c$ by $\delta$ value
22: **end while**
23: **return** $(\theta^*, E_l, Q_l)$

The inputs to the XAI-Gen algorithm are the total number of layers $L$ and the total number of neurons $Y_l$ in a layer $l$ of a pre-trained model $\theta$, a set of approximate multipliers $m$, step size $\delta$, an initial layer conductance threshold $tc$, skipping threshold $tp$, quantization scale $q$ and dataset $\mathcal{D}$. XAI-Gen also takes energy constraint $E_c$, quality constraint $Q_c$, and error magnitude of approximate multipliers $M_g$ (mean average error) as input. The quality constraint refers to the desired classification accuracy defined by the user. To elaborate, XAI-Gen first initializes an array $R$ of accurate multipliers, which will be replaced with the most suitable approximate multipliers for each layer in the subsequent steps (Line 1). Then, XAI-Gen estimates the neuron conductance of each neuron in a pre-trained model $\theta$ (Lines 4-5) and aggregates the neuron conductance in each layer to find the layer importance (Line 7). The layers are identified in order of their highest to lowest importance (Line 9). While traversing through the layers in descending order of importance, XAI-Gen assigns the least approximate multipliers $m_l^*$ to the most important layer and updates $R^*$ with those approximate multipliers (Line 11). The degree of approximation needed for a layer $l$ is determined based on the threshold $t_c$ on the layer importance. The approximate multipliers are selected from a library of approximate multipliers, such as Evoapprox8b [15]. Next, XAI-Gen identifies the indices of the least important neurons $n_p$ in the pre-trained model $\theta$ (Line 12) and skips only $t_p$ number of those neurons (Line 13). After this, XAI-Gen quantizes the model (Line 14) to integrate the $q$-bit approximate multipliers. XAI-Gen uses the updated array $R^*$ to build the AxDNN $\theta^*$ from the quantized model $\theta_q$ (Line 12). XAI-Gen evaluates the energy and quality of the AxDNN, using the dataset $\mathcal{D}$ and analytical model of the multi-pod approximate systolic array (as discussed in Section III), under user-defined energy and quality constraints (Line 13). It is worth mentioning that XAI-Gen uses the *pre-computed* energy values of MAC operations (through Synopsys design compiler) in the analytical model of the hardware accelerator to estimate the energy consumption of the whole AxDNN. The energy of each operation is calculated offline using hardware design tools such as Synopsys Design Compiler. Running XAI-Gen simulations without integrating such tools in the design flow helps in architectural model flexibility and fast hardware efficiency estimation through the analytical model. If the estimated energy consumption and quality meet the energy and quality constraints, then the $t_c$ and the selected approximate multiplier are committed in a layer. In this paper, the simulations are stopped when this evaluation criterion is not satisfied (Lines 14-17). They can also be run to analyze the accuracy-energy tradeoffs for all possible configurations of approximate multipliers in AxDNN layers. XAI-Gen reduces the $t_c$ value by the amount $\delta$ for the subsequent layers (Line 18). The outcome of this algorithm is the AxDNN model $\theta^*$, quality, and energy consumption.

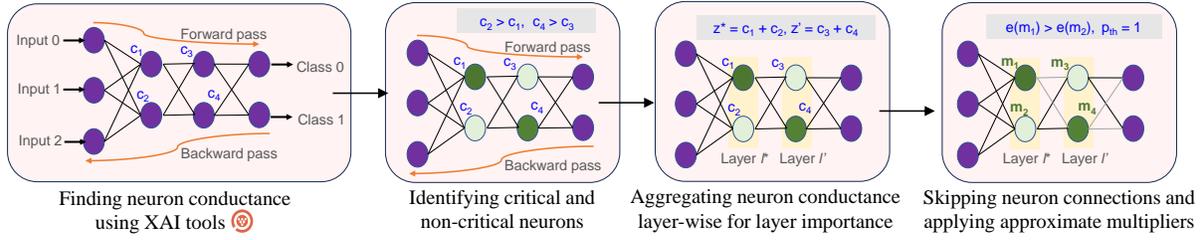

Fig. 3: XAI-guided approximate computing: Skipping the non-critical neurons and applying approximate multipliers layer-wise by identifying the neuron conductance and layer importance using the XAI tools

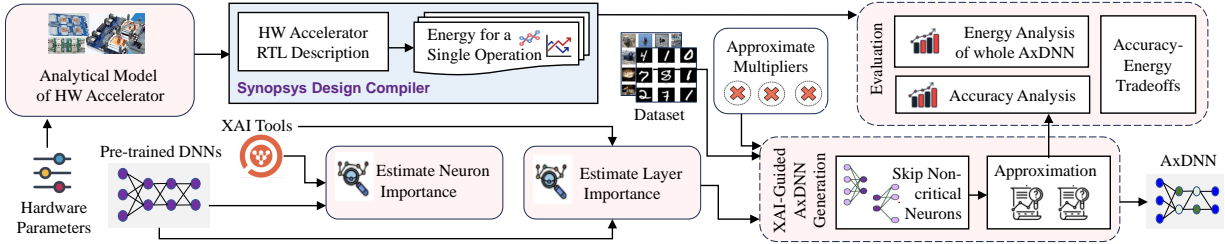

Fig. 4: XAI-Gen: Proposed methodology for generating AxDNNs using XAI techniques

### C. Application of XAI-guided Neural Architecture Search

NAS is a computationally intensive approach for identifying optimized neural networks and AxDNN architectures with the most suitable *homogeneous* approximate multiplier across all layers. This paper uses evolutionary NAS [5] based on Non-dominated Sorting Genetic Algorithm (NSGA-II) as a case study for the XAI-Gen algorithm. As illustrated in Fig. 5. XAI-NAS utilizes the protobuf specification of quantized AxDNN architectures and replaces quantized convolutional layers (QConv2D) with approximate layers containing approximate multipliers using the transform graph tool. XAI-NAS selects heterogeneous approximate multipliers for AxDNNs following the XAI-Gen algorithm, which helps focus on skipping non-critical neurons and applying approximate multipliers to non-critical layers through adaptive layer conductance thresholds, thus reducing search complexity by optimizing for energy-efficient architectures. The search starts with 36 AxDNNs with varying skipping thresholds. Each iteration selects 50 best solutions and generates 50 new candidate AxDNNs with optimal multipliers. The mutation probability is 10%. During XAI-NAS, inference operations use quantized and approximate multiplications, but initial training uses standard floating-point multiplications. To avoid time-consuming retraining, AxDNN weights are tuned using a pre-computed offline weight mapping function [5] for each approximate multiplier, restoring accuracy without retraining. The result of XAI-NAS is a set of optimized AxDNNs that offer higher energy efficiency with minimal accuracy loss.

## V. RESULTS AND DISCUSSIONS

This section details the experimental setup, including datasets, architectures, and parameters, followed by a compre-hensive analysis of the results obtained from evaluating XAI-Gen methodology and its application to NAS (XAI-NAS).

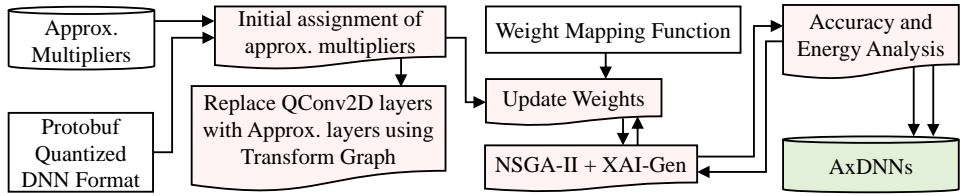

Fig. 5: XAI-NAS: Explainable Neural Architecture Search

### A. Datasets and Architectures

We use three datasets, i.e., MNIST, CIFAR10, and ImageNet. The DNN research community widely uses these datasets for evaluating the performance of neural networks [3], [24], [25]. We use 80% of the data samples for training and the remaining 20% samples to evaluate AxDNNs in the inference phase. Note that the test server of ImageNet 2012 is no longer available; we use the accuracy on the validation set as the test accuracy. For MNIST [13], we train Lenet5 with a baseline accuracy of 98%. For ImageNet, we train ResNet-50 with a baseline accuracy of 90%, which is consistent with reported results for 8-bit quantized model. For CIFAR-10 [14], we train ResNet-18 and Resnet-34 with a baseline accuracy of 91% and 92%, respectively. The XAI-Gen considers eight $256 \times 256$ pods for the multi-pod systolic array accelerator.

### B. Parameters for Approximation Analysis

During the approximation analysis in Algorithm 1, the AxDNN models are implemented using the state-of-the-art AdaPT [24] library. Each AxDNN employs 8-bit Pareto-optimal approximate multipliers selected based on their error magnitude $M_g$ from the EvoApprox8b [15] library. The $M_g$ refers to the standard error metric, i.e., mean average error (MAE) for approximate multipliers. The layer importance of each AxDNN is estimated using the Captum [21]. The $Q_c$ is set as 2% below the baseline accuracies of LeNet-5, ResNet-18, ResNet-34, and ResNet-50. The $\mathcal{R}$ is set as 128 for experiments. The energy consumed by the MAC units is estimated using the Synopsys Design Compiler. The energy per memory access is calculated as 2.7 pJ/Byte using the Cacti-P tool [26].

### C. XAI-Guided Approximation Analysis

This section discusses the layer importance, accuracy-energy tradeoffs of XAI-Gen generated AxDNN, and execution time of XAI-Gen powered XAI-NAS case study.

**Analyzing layer importance:** To evaluate XAI-Gen's effectiveness in identifying layer importance, we conduct five experiments per AxDNN for different target classes, which guide our predictions. Unrolling and analyzing each layer of complex AxDNNs with Captum can be computationally intensive, especially for larger ResNet architectures. Consequently, we aggregate conductance values for all layers within a residual block to assess block importance. As shown in Fig. 6a, the second convolutional layer of LeNet-5 is relatively more important, whereas Fig. 6b, Fig. 7a, and Fig. 7b reveal that in ResNet-18, ResNet-34, and ResNet-50, the first convolutional layer is most crucial, with importance decreasing from input to output. This suggests XAI-Gen assigns more significance to input layers and applies less or no approximation to them, aligning with state-of-the-art work [2]. Nonetheless, discrepancies may arise due to varying neuron vulnerability to approximation errors based on the output class.

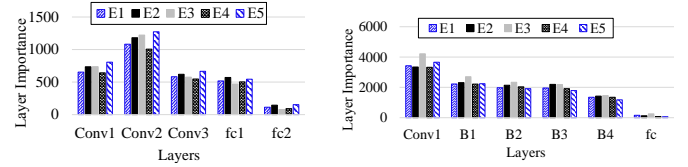

(a) LeNet5 architecture     (b) ResNet-18 architecture

Fig. 6: Layer importance of convolutional and fully connected layers in LeNet-5 and the residual blocks (B$m$) of ResNet-18. $E_n$ to $E5$ represent five experiments for attributing predictions with different target classes

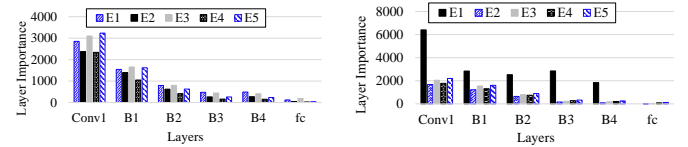

(a) ResNet-34 architecture     (b) ResNet-50 architecture

Fig. 7: Layer importance of convolutional and fully connected layers, and residual blocks (B$m$) in ResNet-34 and ResNet-50. $E1$ to $E5$ represent five experiments for attributing predictions with different target classes

**Analyzing accuracy and energy tradeoffs:** We further investigate the performance of XAI-Gen by analyzing the accuracy-energy tradeoffs of different AxDNNs. Since we calculated the aggregated importance of the layers in a residual block, we applied the same approximate multiplier to all residual layers within the block. This helps reduce the exhaustive computational space of the most suitable approximate multiplier. Note that the heterogeneity of the approximate multipliers is maintained in the other AxDNN layers. The energy consumed by the accurate LeNet5, ResNet18, ResNet-34, and ResNet-50 architectures ($A$) is 15, 45, 60, and 70 microwatts ($\mu W$), respectively. Figs. 8 and 9 illustrate the energy consumption and accuracy tradeoffs for different AxDNN configurations. The most energy-efficient configurations, achieved by skipping non-critical neurons ($p$) followed by approximation ($a$), result in energy consumptions of 2, 17, 34, and 40 microwatts for approximate LeNet5 ($A2 + pa$ in Fig. 8a), ResNet18 ($A5 + pa$ in Fig. 8b), ResNet-34 ($A6 + pa$ in Fig. 9a), and ResNet-50 ($A3 + pa$ in Fig. 9b), respectively. **Thus, XAI-Gen leads to almost $7\times$, $2.6\times$, $2\times$, and $2\times$ lower energy consumption** in approximate LeNet5, ResNet18, ResNet-34, and ResNet-50, respectively. *Interestingly, some approximate architectures achieve up to 6% higher accuracy with approximation*. For example, $A6+pa$ and $A5 + pa$ in the case of approximate LeNet5 and ResNet-18. This is because XAI assigns the importance score to the neurons according to the target classes. When we remove the non-critical neurons that do not significantly impact the image classification, the accuracy of AxDNNs is slightly improved in some cases. Table I presents the most suitable configurations of skipping thresholds and approximate multipliers for AxDNN architectures generated by the XAI-Gen algorithm. It is worth noticing that despite 70% of the skipped neurons and approximate multiplier with the highest approximation error (0.52%) in a layer, the AxDNN architectures generated by XAI-Gen only suffer from **1%-2% accuracy loss only**. This shows that XAI-Gen carefully skips the non-critical neurons and applies approximation to the non-critical layers.

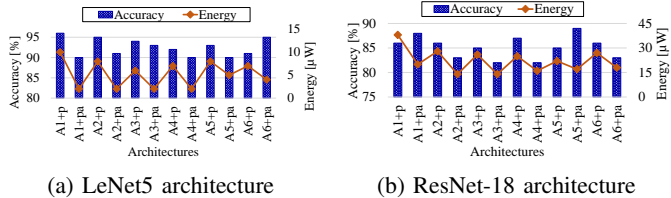

(a) LeNet5 architecture  (b) ResNet-18 architecture

Fig. 8: Accuracy-energy tradeoffs for LeNet5 and ResNet-18 architecture (A) with XAI-guided skipping of non-critical neurons (p) and approximation (pa)

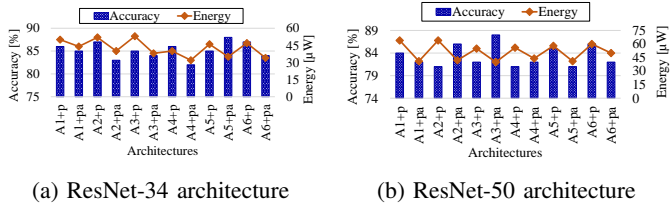

(a) ResNet-34 architecture  (b) ResNet-50 architecture

Fig. 9: Accuracy-energy tradeoffs for ResNet-34 and ResNet-50 architectures (A) with XAI-guided skipping of non-critical neurons (p) and approx. (pa)

### D. XAI-NAS: XAI-guided Neural Architecture Search

Since XAI-Gen adapts the layer conductance threshold for an appropriate selection of the appropriate approximate multipliers instead of an exhaustive hit-and-trial method, integrating XAI-Gen in NAS can expedite the search process. XAI-NAS takes advantage of XAI-Gen to incorporate *heterogenous* approximate multipliers in the AxDNN layers. This helps achieve higher energy efficiency than NAS. Fig. 10 shows different generations (g) of the AxDNN search process for the CIFAR10 dataset. The earliest generations contain sub-optimal AxDNNs. However, most of the Pareto-optimal AxDNNs are observed in the latest generations. Interestingly, we observe that XAI NAS for ResNet-14 in Table II achieves almost 7% and 10% improvement in the accuracy when compared to the existing NAS approaches like ALWANN [5] and CGP [6], respectively. *This improvement is attributed to XAI, as discussed in the previous section.* Compared to the baseline accuracies, we observe that XAI-NAS achieves up to $2.5\times$ and $2\times$ lower energy consumption when compared to ALWANN and CGP, respectively, while maintaining comparable or better accuracy. Note the relative accuracy of ResNet-14 is 99%. Furthermore, we also observe that the XAI-NAS results in up to **40% higher energy efficiency** even with 1% difference in accuracy due a change in the approximate multiplier configuration in approximate ResNet architectures with the CIFAR10 dataset. This is because XAI-NAS, using XAI-Gen, only applies approximation to non-critical neurons and layers. On the other hand, the NAS achieves up to only 30% higher energy efficiency with 1.7% accuracy loss. *This means that integrating our proposed XAI-Gen with the NAS algorithm helps achieve up to 10% higher energy efficiency when compared to the state-of-the-art NAS algorithm.*

Table I: Best parameter configurations for approximate LeNet5 (L5), ResNet18 (R1), ResNet-34 (R2), and ResNet-50 (R3) architectures generated with XAI-Gen. M1 is an accurate multiplier. M2-M7 refers to approximate multipliers namely, KV8 (0.0018% MAE), KV9 (0.0064% MAE), KVP (0.051% MAE), L2J (0.081% MAE), L2L (0.23% MAE), and L2N (0.52% MAE) in Evoapprox8b [15] library.

| | Skipping Threshold | Multipliers | Accuracy | $E_a$ |
|---|---|---|---|---|
| L5 | (5, 0, 10, 15, 70) | (M2, M1, M3, M6, M7) | 94% | 2 $\mu$W |
| | (5, 0, 5, 10, 60) | (M3, M1, M3, M6, M7) | 94% | 3 $\mu$W |
| | (5, 0, 5, 20, 50) | (M2, M1, M2, M6, M7) | 95% | 4 $\mu$W |
| R1 | (0, 5, 5, 10, 15, 70) | (M1, M2, M2, M5, M6, M7) | 88% | 26 $\mu$W |
| | (0, 5, 5, 10, 20, 70) | (M1, M2, M2, M3, M4, M5) | 86% | 20 $\mu$W |
| | (0, 5, 10, 15, 15, 60) | (M4, M1, M5, M6, M7) | 87% | 17 $\mu$W |
| R2 | (0, 5, 10, 15, 15, 50) | (M1, M2, M2, M3, M3, M7) | 81% | 38 $\mu$W |
| | (0, 5, 10, 15, 15, 70) | (M1, M2, M2, M4, M4, M5) | 89% | 35 $\mu$W |
| | (0, 5, 10, 30, 35, 75) | (M1, M2, M3, M6, M7) | 88% | 34 $\mu$W |
| R3 | (0, 10, 10, 15, 25, 55) | (M1, M2, M2, M3, M6, M7) | 91% | 42 $\mu$W |
| | (0, 15, 20, 25, 30, 60) | (M1, M2, M3, M4, M5, M6) | 89% | 40 $\mu$W |
| | (0, 15, 30, 30, 35, 35) | (M1, M2, M3, M3, M6, M7) | 90% | 41 $\mu$W |

Table II: Comparing accuracy and energy efficiency of ALWANN [5] and CGP [6] with our XAI-NAS for ResNet-14

| Perf. Metrics | ALWANN [5] | CGP [6] | XAI-NAS |
|---|---|---|---|
| Accuracy | 85.55% | 83.98% | 92% |
| Energy | 19.76 $\mu$J | 14.88 $\mu$J | 7.88 $\mu$J |

Since we leverage XAI techniques for accelerated adoption of the appropriate heterogeneous approximate multipliers in AxDNN layers, the evaluation of our approximate ResNet-50 generated with XAI-Gen takes only 350 seconds, unlike the manual search of 310 GPU hours as highlighted in Section I-A. Such a huge reduction in execution time makes XAI-NAS very quick in searching the optimized AxDNN architectures. Table III compares the execution time of our XAI-NAS for finding the most suitable approximate multipliers for different ResNet architectures with the existing NAS algorithms. The CGP-based NAS [6] takes 3.2 seconds in the case of ResNet-8 architecture with the CIFAR-10 dataset. However, our XAI-NAS takes only 1.2 seconds. This means that despite our XAI-NAS considering heterogeneous approximate multipliers across AxDNN layers, it is up to $\approx 3\times$ faster than CGP-based NAS. Interestingly, ALWANN [5] takes 5.14 GPU hours in the search and validation process with the CIFAR10 dataset; however, our XAI-NAS takes only 1 hour when simulated on the same GPU. This means that despite our XAI-NAS, it is **up to $5\times$ faster than ALWANN**. A similar trend is also observed for other ResNet architectures.

Table III: Comparing execution time of the ALWANN [5] and CGP [6] with our XAI-NAS for AxDNNs

| AxDNNs | ALWANN [5] | CGP [6] | XAI-NAS |
| --- | --- | --- | --- |
| ResNet-8 | 28.8 min | 3.2 sec. | 1.2 sec |
| ResNet-14 | 1.58 hours | 45 sec. | 15 sec. |
| ResNet-50 | 5.14 hours | 3 hours | 1 hour |

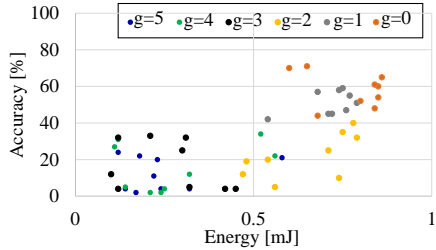

Fig. 10: XAI-NAS generations (g) for CIFAR-10 dataset

## VI. CONCLUSION

Approximate computing has great potential to improve the energy efficiency of edge analytics through the application of approximate multipliers in AxDNNs. However, searching for appropriate approximate multipliers layer-wise for better energy-accuracy tradeoffs AxDNNs can be very time-consuming. This paper addresses this challenge through a novel XAI-Gen methodology that harnesses explainable artificial intelligence (XAI) to generate AxDNNs. XAI-Gen aims to provide higher energy efficiency at the cost of minimal accuracy loss in AxDNNs on multi-pod systolic array accelerators. Our results show that XAI-Gen achieves up to $7\times$ lower energy consumption with up to 1-2% accuracy loss. We also show that XAI-Gen is very effective in the neural architecture search (XAI-NAS) for AxDNNs. XAI-NAS achieves 40% higher energy efficiency with $5\times$ less execution time compared to other state-of-the-art NAS for AxDNNs. To the best of our knowledge, this is the first work leveraging XAI for fast AxDNN generation and NAS, opening new avenues for efficient approximate computing in deep learning. Future work will focus on extending our proposed approach to more advanced DL architectures, such as Transformer models, to explore its scalability and effectiveness in diverse neural network paradigms. We also plan to adapt our method for other types of AI accelerators like Eyeriss and Simba, aiming to demonstrate its versatility across different hardware platforms.